%% file: PaperForReview.tex
\documentclass[10pt,twocolumn,letterpaper]{article}

\usepackage[pagenumbers]{cvpr} 
\usepackage{graphicx}
\usepackage{amsmath}
\usepackage{amssymb}
\usepackage{booktabs}
\usepackage{multirow}
\usepackage{hhline}
\usepackage{booktabs}
\usepackage[table,xcdraw]{xcolor}
\usepackage{svg}

\usepackage[pagebackref,breaklinks,colorlinks]{hyperref}

\usepackage{pifont}%
\newcommand{\cmark}{\ding{51}}%
\newcommand{\xmark}{\ding{55}}%
\usepackage[capitalize]{cleveref}
\crefname{section}{Sec.}{Secs.}
\Crefname{section}{Section}{Sections}
\Crefname{table}{Table}{Tables}
\crefname{table}{Tab.}{Tabs.}

\def\cvprPaperID{32} 
\def\confName{CVPR}
\def\confYear{2023}

\newif\ifprintcomments
\printcommentstrue

\begin{document}

\title{Just a Glimpse: Rethinking Temporal Information for Video Continual Learning}

\author{Lama Alssum, Juan León Alcázar, Merey Ramazanova, Chen Zhao, Bernard Ghanem\\
King Abdullah University of Science and Technology (KAUST)\\
{\tt\small \{lama.alssum.1, juancarlo.alcazar, merey.ramazanova, chen.zhao, bernard.ghanem\}@kaust.edu.sa}
}
\maketitle

\input{sections/abstract}
\input{sections/introduction}

\input{sections/relatedwork}

\input{sections/methodology}

\input{sections/experiments}

\input{sections/conclusion}

{\small
\bibliographystyle{ieee_fullname}
\bibliography{egbib}
}

\end{document}

%% file: sections/abstract.tex
\begin{abstract}
Class-incremental learning is one of the most important settings for the study of Continual Learning, as it closely resembles real-world application scenarios. With constrained memory sizes, catastrophic forgetting arises as the number of classes/tasks increases.
Studying continual learning in the video domain poses even more  challenges, as video data contains a large number of frames, which places a higher burden on the replay memory. The current common practice is to sub-sample frames from the video stream and store them in the replay memory. In this paper, we propose SMILE a novel replay mechanism for effective video continual learning based on individual/single frames. Through extensive experimentation, we show that under extreme memory constraints, video diversity plays a more significant role than temporal information. Therefore, our method focuses on learning from a small number of frames that represent a large number of unique videos.
On three representative video datasets, Kinetics, UCF101, and ActivityNet, the proposed method achieves state-of-the-art performance, outperforming the previous
state-of-the-art by up to 21.49\%.

\end{abstract}

%% file: sections/introduction.tex
\section{Introduction}
\label{sec:introduction}

Recently, a large amount of data has become available on the Internet due to high-speed internet access and the rapid growth of social media platforms. In particular, sharing and uploading video data has become popular with the availability of video cameras in mobile and wearable devices. As the number of available videos grows, so does the required human labor to analyze them. This expensive manual labor motivates the development of intelligent systems for video understanding and analysis. Recently, a growing interest of the research community in human action recognition has led to the development of massive video datasets and large-scale models with a large number of action classes \cite{carreira2017quo, karpathy2014large, tran2015learning, wang2018temporal}.

Most of the previous work in video action recognition assumes that a large amount of labeled data from a predefined number of classes is available, and this class set remains fixed once the model is deployed. These assumptions do not always hold in real-world applications where novel classes are continually identified, and collecting enough labeled data is expensive and time-consuming. With a continuous stream of novel data and classes, the development of models that can learn sequentially from a set of tasks is needed. Such a setup requires the sequential fine-tuning of the model on a set of tasks. Under these circumstances, it has been shown that Neural Networks suffer from a phenomenon known as catastrophic forgetting \cite{french1999catastrophic, goodfellow2013empirical, mccloskey1989catastrophic}, where fine-tuning the model on a new task reduces its performance drastically on previously learned tasks. Continual Learning (CL) is the field of study that addresses the challenging setup of training a model on a set of sequential tasks while mitigating catastrophic forgetting\cite{dhar2019learning}.

\input{figures/bubble}

We focus on a special case of continual learning called Class Incremental Learning (CIL) \cite{masana2020class}, where the model learns on a set of tasks that contain disjoint classes with access to a limited amount of previously seen data, named experience replay (ER) \cite{rolnick2019experience}. Currently, ER methods have shown superior performance in the CIL setup \cite{Lomonaco2022CVPR2C, Merlin2022PracticalRF} compared to their regularization-based counterparts \cite{aljundi2018memory, kirkpatrick2017overcoming, zhang2020class, chaudhry2018riemannian}. 

Previous work has already explored the CIL setup in the video domain \cite{park2021class, villa2022vclimb, ma2021class, zhao2021video, pei2022learning}. However, video data requires a significantly larger memory size in comparison to image data; due to the additional temporal dimension. This drawback is a central motivation for the development of video CL strategies endowed with an efficient working memory management strategy. Despite the clear relevance of memory efficiency  in video CL, it remains an under-studied topic. 

In this work, we set our attention towards one of the main limitations of ER-based methods in the video continual learning setup, namely the memory constraint. ER methods suffer from catastrophic forgetting, as the limited memory capacity allows only a fraction of previously seen data to be stored, and the in-memory training subset typically diverges from the original distribution of the training data \cite{clear,cloc,aljundi2017expert,li2017learning}. Moreover, as the task number increases, some elements must be evicted from the memory buffer to accommodate examples of novel classes. The model is then updated with increasingly skewed versions of the original distribution. 

We depart from the golden standard in video continual learning, as we no longer attempt to retain any temporal information in the memory buffer. In fact, we propose an extreme sampling strategy that completely ignores temporal data in favor of greedily sampling individual frames from as many training videos as possible. We empirically show that our extreme sampling strategy, paired with an inexpensive training time regularization, can outperform the well-established ER strategies that focus on preserving the temporal progression of the video data, as shown in Figure \ref{figure:bubble}, while also reducing the total amount of required memory. Moreover, since we store a single frame from every training sample, we create a direct equivalence between the memory buffers devised for image continual learning and those for video continual learning. This enables the direct application of image CIL methods for video CIL. 
This paper introduces SMILE (Single fraMe Incremental LEarner) for video CIL. Our work brings the following contributions:

\begin{itemize}
    
    \item We introduce an extreme memory sampling strategy for video CIL, where a single representative frame is stored in memory. Paired with our regularization technique, this sampling can reach state-of-the-art performance across all datasets in the vClimb benchmark while using far less replay memory.
       
    \item We show that storing a single frame in memory allows the direct use of image-domain CL methods for video data. This direct adaptation outperforms all other CIL methods in the challenging vClimb benchmark.
\end{itemize}

%% file: figures/bubble.tex
\begin{figure}[t!]
\begin{center}
\includegraphics[width=0.46\textwidth]{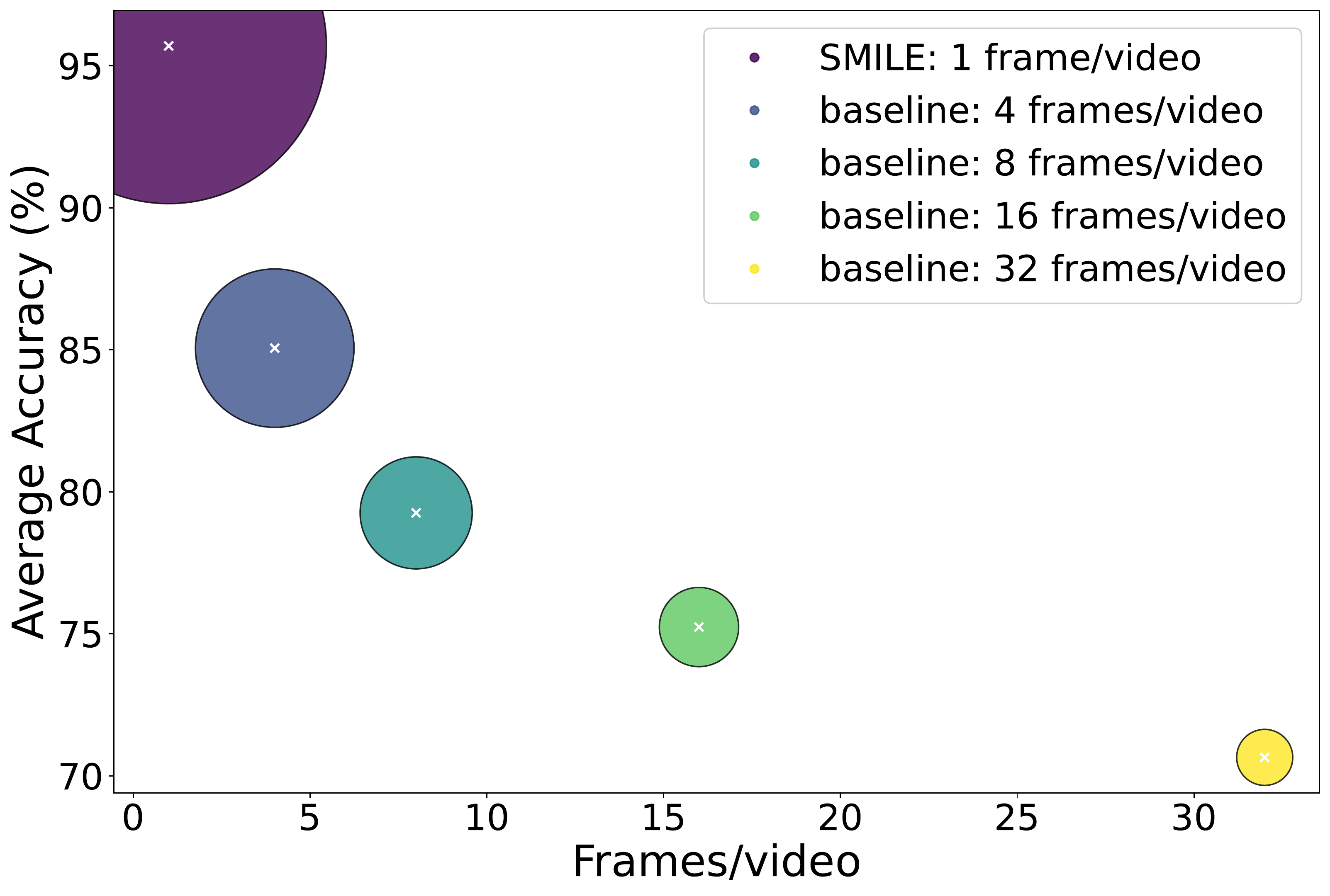}
\small
\caption{\textbf{SMILE \vs the baseline performance (iCaRL)}. Although our method (SMILE) is restricted to storing a single frame per video, this memory-efficient sampling outperforms the scenarios that focus on preserving the temporal progression of the video data in the UCF101 dataset, regardless of the temporal resolution of the videos stored in the memory buffer. In a restricted memory setting (\eg, rehearsal methods for continual learning), video diversity (represented by the disc size) is the most crucial factor in achieving high performance in video class incremental learning.}

\label{figure:bubble}
\end{center}
 \vspace{-15pt}
\end{figure}

%% file: sections/relatedwork.tex
\section{Related Work}
\label{sec:relatedwork}

Most of the literature approaches the CL problem in the image domain. For image classification, current methods can be divided into three sub-categories \cite{de2021continual}: regularization, replay memory, and parameter isolation. 
\input{figures/pull.tex}

\textit{ i) Regularization methods} aim at building and retaining a parameter-set that can perform consistently across tasks \cite{aljundi2018memory,kirkpatrick2017overcoming,zhang2020class,li2017learning}. That is, for every individual task, a subset of the most relevant network weights is identified, and a penalty is applied for updating these parameters. It is expected that the critical parameter sub-sets for individual tasks are kept partially stable across the continual learning process.

\textit{ ii) Rehearsal methods} mitigate catastrophic forgetting by storing a subset of the original training samples \cite{rebuffi2017icarl,wu2019large,lopez2017gradient,prabhu2020gdumb, hou2019learning} or by fitting a generative model to training data of previous tasks \cite{hayes2020remind,lesort2019generative,shin2017continual}, thus enabling the generation and storage of surrogate training samples. This memory is replayed while learning a novel task, typically by using it as additional training data or by estimating boundary points that guide the optimization process on novel tasks. 

\textit{ iii) Parameter isolation methods} tackle catastrophic forgetting by segmenting and specializing the parameter space in the Neural Network \cite{mallya2018piggyback,rusu2016progressive,serra2018overcoming}. These models generate parameter sub-spaces specialized on an individual task, thus preventing performance loss across tasks. In its most simple way, they create alternative branches for data processing inside the network while freezing the existing parameter set from previous tasks \cite{hu2019overcoming,xu2018reinforced}.

\vspace{3pt}\noindent\textbf{Video Continual Learning.} Recently, some works have started approaching the Continual Learning problem in the video domain \cite{park2021class,villa2022vclimb, ma2021class, zhao2021video}. Notably, the release of the vCLIMB benchmark \cite{villa2022vclimb} has provided the first standard test-bed for video CIL, along with the initial baseline results of classic CIL methods \cite{rebuffi2017icarl,wu2019large} applied into video data. 

Currently, the best-performing methods in vCLIMB are rehearsal approaches where a subset of the video frames is stored in the replay memory. Such an approach suffers from two drawbacks: First, methods must sub-sample the frames to be stored as pushing a full video in the replay memory is prohibitively expensive. Second, video data might contain some background frames which are not relevant to the classification task. However, memory methods still lack a frame selection strategy that could directly select the best frames to store in memory. Third, it is still unknown how to deal with the temporal dimension of videos; in other words, how to sample a memory that retains the relevant temporal cues contained in the original video.

To mitigate these drawbacks, we follow a rehearsal strategy but radically depart from the standard approach of storing a sub-set of frames in memory. Instead, we explore the extreme scenario where a single frame is stored in the replay memory. We show that without any bells and whistles, this extreme memory setup can outperform state-of-the-art methods in the vClimb benchmark.

\vspace{3pt}\noindent\textbf{Rehearsal Methods for Video Action Recognition.}  Following up with the previous line of research on the continual learning in videos \cite{villa2022vclimb, park2021class, zhao2021video,ma2021class}, we benchmark the performance on the task of action recognition. Previous works used frame-based approaches for the video representation methods \cite{wang2018temporal, lin2019tsm}. In this setup, several frames are sampled from temporal segments in a video and stored in the memory. In our work, we show how storing in the memory as few as one frame per video can result in a better trade-off, as more diverse video information could be preserved then.

%% file: figures/pull.tex
\begin{figure*}[t]
\begin{center}
\includegraphics[width=\textwidth]{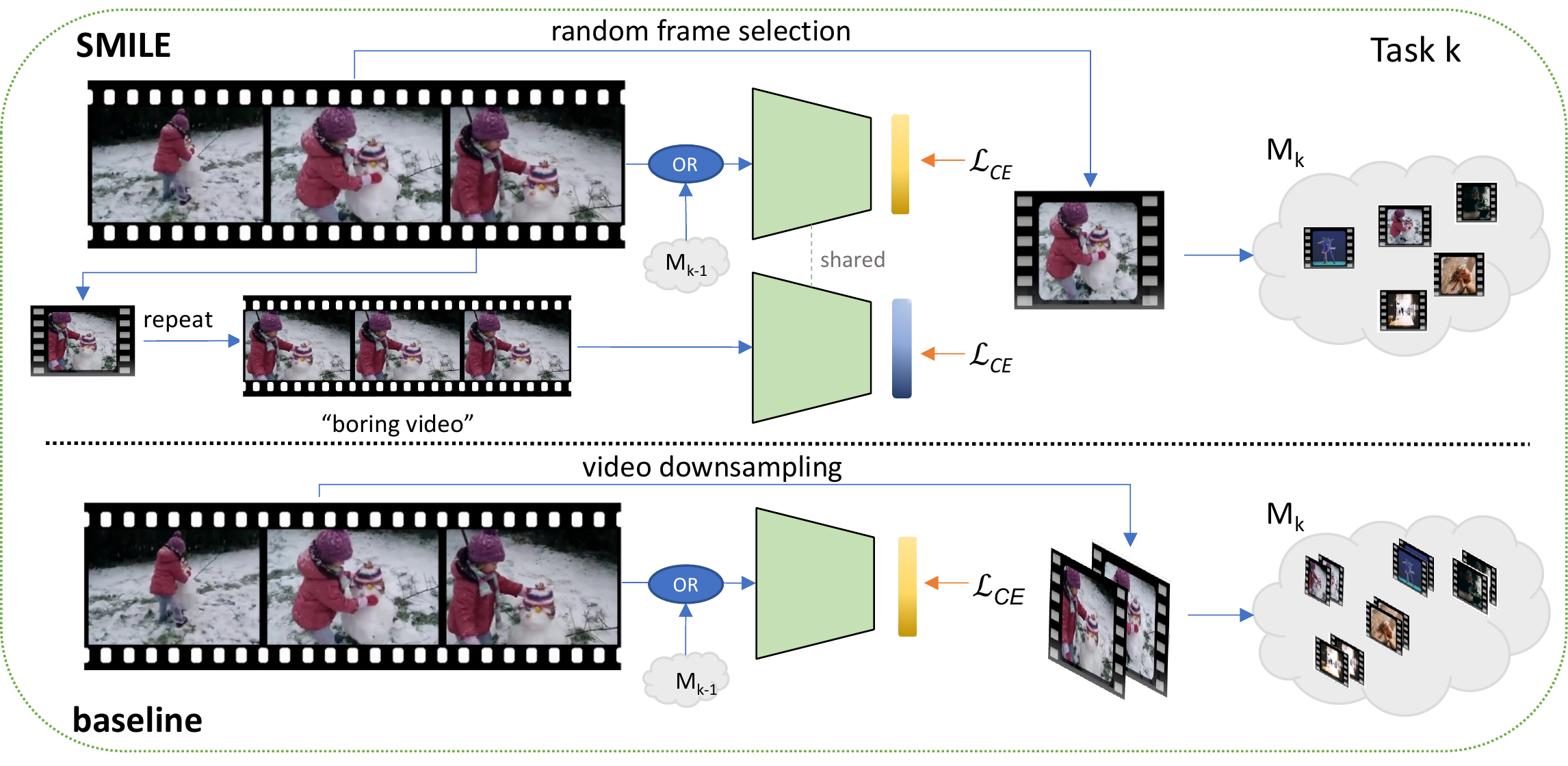}
\small
\caption{\textbf{SMILE \vs baseline methods.} SMILE relies on sampling a single frame per video into the replay memory. This sampling favors video diversity and improves the performance in the CIL setup. In addition, it drastically reduces memory usage. However, it removes the temporal information from memory. We develop a secondary forward pass that helps the video encoder to adapt to both modalities (video and static images). Under this sampling strategy, we can directly use image-based continual learning methods in the video domain.  }
\label{figure:pull}
\end{center}
 \vspace{-10pt}
\end{figure*}

%% file: sections/methodology.tex
\section{SMILE:  \underline{S}ingle Fra\underline{m}e \underline{I}ncremental  \underline{Le}arner}
\label{sec:methodology}

In this section, we outline the details of our proposed approach. Since we rely on single-frame memories, we name our method SMILE (\textbf{S}ingle fra\textbf{M}e \textbf{I}ncremental \textbf{LE}arner). Before delving into the details of SMILE, we first provide a formal definition of the video class incremental learning (CIL) problem and then motivate our extreme memory design with an initial experiment on the UCF101 dataset~\cite{soomro2012ucf101}. 

\noindent\textbf{Notation and Problem Formulation.} In class incremental learning, a neural network, $f_\theta : \mathcal  X \rightarrow \mathcal Y$  parameterized by $\theta$, is trained on a set of $m$ sequential tasks $T=\{t_1, t_2,.., t_k, ..,t_m\}$.  Each $t_k$ is composed of $n_k$ samples $\{(\mathbf{x}_1^k,y_1^k), (\mathbf{x}_2^k,y_2^k), ..,(\mathbf{x}_{n_{k}}^k,y_{n_{k}}^k)\} \sim \mathcal{D}_k$ where 
$\mathbf{x}_i^k$ represents a training sample and $y_i^k$ is its corresponding ground truth label in the dataset $\mathcal{D}_k$. The classes of samples introduced at each task $\left\{y_i^k\right\}_{i=1}^{n_k} \in \mathcal{Y}_k$ are unique, \ie $\bigcap_{i=1}^m \mathcal{Y}_i = \emptyset$.

\subsection{Analyzing Replay Memories for Video CIL}

We motivate the single frame sampling by summarizing the memory constraints defined in vCLIMB~\cite{villa2022vclimb} and explaining the trade-off between dense temporal sampling and memory diversity. We define memory diversity as the number of unique video clip samples stored in the memory. 

The vCLIMB Benchmark defines the replay memory in terms of frames. That is, a video must be stored as a full set of contiguous frames or as a temporally sub-sampled version of it. 
This design choice of using the number of frames to define the memory size enforces a common ground for comparison between methods, regardless of the length of the video data. Moreover, it creates a trade-off unique to video CIL, we must choose whether to store full videos (fewer unique videos, with full temporal data) or store sub-sampled videos (more unique videos, incomplete temporal data). 

\input{Tables/Analysis}

In Table \ref{table:analysis}, we provide the initial insight into our approach. We set a fixed memory budget of 16160 frames and analyze the effect of pushing 4, 8, 16, or 32 frames per video into the replay memory. We also report the number of unique videos stored in memory after the final task. We deploy a CIL setup on UCF101 \cite{soomro2012ucf101} using iCaRL \cite{rebuffi2017icarl}. This simple analysis suggests that experience replay (ER) methods might benefit from a regime where the replay memory is filled with less temporal data (4 frames per video) while favoring video diversity (4040 unique videos in memory), instead of a regime with more temporal information (32 frames per video) and less video diversity (505 unique videos). 

Based on these initial observations, we propose to improve the working memory usage by maximizing the diversity of samples contained in it. We focus on the extreme case of sub-sampling, where we maximize the diversity of in-memory elements by sampling a single frame into the working memory. Such an extreme sampling strategy represents a two-fold improvement as follows. First, it drastically reduces the total amount of required memory. Second, it enables the direct application of ER methods from the image domain. However, by sampling a single frame, we will lose the temporal dimension for the samples stored in the working memory. To mitigate the forgetting associated with the lack of temporal information in the memory bank, we develop a novel training-time regularization detailed in the following section.  

\subsection{Single Frame Memories for Video CIL}

Following our memory sampling strategy, two types of data will be available during the CIL process. For every new task, video clips with full temporal resolution will be available. Meanwhile, the data from previous tasks will be represented exclusively by images (single frame per video) stored in the replay memory. To adapt to these two data sources, we begin by ``inflating'' the input tensor for image data. We follow \cite{carreira2017quo} to expand the frame data stored in memory into a ``boring video'', namely, a video clip composed of $n$  copies of the original frame. 

Boring videos enable us to learn from both types of data with the same video encoder, however, these boring videos introduce a  distribution shift between real video clips and in-memory images (represented as boring videos). To address this undesired shift, we propose the following loss:
\begin{equation*}
    \mathcal{L} = \lambda_{1} \mathcal{L}_{ce}(X_{k},Y_{k}) +  \lambda_{2} \mathcal{L}_{ce}(\hat{X}_{k},Y_{k}) + \lambda_{3} \mathcal{L}_{ce} (X_{m}, Y_{m}),
\end{equation*}
\noindent where $\mathcal{L}_{ce}$ is cross-entropy loss. $X_{k}$ and $\hat{X}_{k}$ are the video samples of the task $t_{k}$ and the boring video version of them, respectively. $Y_{k}$ is the ground truth labels for $X_{k}$. $X_{m}$ and $Y_{m}$ are the samples stored in memory and their ground truth labels. $\lambda_{1}$, $\lambda_{2}$, and $\lambda_{3}$  are weights to control the contribution of each loss term. The secondary loss ($\mathcal{L}_{ce}(\hat{X}_{k},Y_{k})$) directly addresses the domain shift, as it allows the video encoder to slowly adapt to the two data sources (standard and boring video) as the CIL tasks progress.

In every new task, we optimize our model using the original video samples (available as the task training data), the boring video version of them, and the boring videos from previous tasks (stored as images in replay memory). Since the boring and the original videos can be forwarded through the same encoder, we simply gather the samples after the forward pass and re-group them in the appropriate factor of the loss function. The pipeline of our approach is illustrated in Figure \ref{figure:pull}.

\vspace{3pt}\noindent\textbf{CIL Methods.} Our single-frame working memory creates a direct link between the image domain and the video domain. We choose two representative methods from the image domain, BiC \cite{wu2019large} and iCaRL \cite{rebuffi2017icarl}. We directly apply both methods to our single-frame memory approach without any changes.

%% file: Tables/Analysis.tex
\begin{table}[t]
\centering

\renewcommand{\arraystretch}{1.6}
\scalebox{0.856}{
\begin{tabular}{@{}c
>{\columncolor[HTML]{EFEFEF}}c c
>{\columncolor[HTML]{EFEFEF}}c c@{}}
\specialrule{0.8pt}{0pt}{0.5pt}
\textbf{Total Memory Size (Frames)} & \textbf{16160}  & \textbf{16160}  & \textbf{16160}  & \textbf{16160}   \\ 
\hhline{-----} 
\hhline{-----}
Unique Videos in Memory & 4040  & 2020  & 1010  & 505   \\ \hhline{-----}
Stored Frames per Video  & 4     & 8     & 16    & 32    \\ \hhline{-----}
Average Accuracy (Acc)        & \textbf{85.06} & 79.26 & 75.24 & 70.65 \\ \hhline{-----}

\specialrule{0.8pt}{0pt}{0.5pt}
\end{tabular}}
\caption{\textbf{Temporal Resolution \vs Memory Diversity.} We empirically assess that the CIL methods seem to favor video diversity over dense temporal sampling. For a fixed memory budget, the direct application of iCaRL to the UCF101 dataset shows that better results are obtained when pushing fewer frames per video (but more unique videos) into replay memory. }
\label{table:analysis}
\end{table}

%% file: sections/experiments.tex
\section{Experimental Evaluation}
\label{sec:experiments}

\input{Tables/Statistics}
\input{Tables/main_results_10}
In this section, we report the results across all the datasets included in the vCLIMB benchmark: UCF101 \cite{soomro2012ucf101}, ActivityNet \cite{caba2015activitynet}, and Kinetics \cite{carreira2017quo}, and provide ablation results to support our design decisions. 

\vspace{3pt}\noindent\textbf{Datasets.} UCF101 has 13.3K videos from 101 classes. ActivityNet consists of 20k videos from 200 classes and can be used for both trimmed and untrimmed action recognition. In this work, we use the trimmed version, where every frame of the video is part of an action instance. Kinetics is a large-scale dataset that has more than 300K short videos from 400 classes. For every dataset, we train SMILE sequentially on task-sets composed of 10 and 20 tasks as defined in \cite{villa2022vclimb}. Statistics for these tasks-sets are presented in Table \ref{table:4}.     

\vspace{3pt}\noindent\textbf{Evaluation Metrics.} To evaluate the performance of SMILE, we use the Average Accuracy metric (Acc) \cite{lopez2017gradient}, which calculates the mean accuracy of all learned tasks after finishing the training on the sequence of tasks $T = \{t_1, t_2, ..,t_m\}$:
\begin{equation*}
  Acc =  \frac{1}{m} \sum_{i=1}^{m} acc_i,
\end{equation*}
where $m$ is the number of available tasks and $acc_i$ is the accuracy of $t_i \in T$. A higher Acc indicates better performance. We also use Backward Forgetting (BWF) \cite{lopez2017gradient} to measure the impact of learning a new task $t_k$ on the performance of previously learned tasks, it is calculated as:
\begin{equation*}
BWF = \frac{1}{m-1} \sum_{i=1}^{m-1} acc_{i,i} - acc_{m,i},
\end{equation*}
where $m$ is the number of tasks learned, and $acc_{m,i}$ and $acc_{i,i}$ are the accuracy on task $i$ after training on task $m$ and the accuracy on task $i$ after training on task $i$, respectively. A higher value of BWF indicates a larger degree of forgetting.

\vspace{3pt}\noindent\textbf{Implementation Details.} 
Both iCaRL and BiC are trained sequentially on each task following the original implementation outlined in \cite{wu2019large} and \cite{rebuffi2017icarl}. SMILE video encoder is based on TSN \cite{wang2018temporal} with a ResNet-34 backbone \cite{he2016deep}, pre-trained with ImageNet weights. We operate on 8 segments per video (\ie we sample $N=8$ frames across evenly spaced temporal segments on a video). We use the SGD optimizer with a batch size of 80 and a learning rate of $1 \times 10^{-3}$ for UCF101 and $6 \times 10^{-4}$ for both ActivityNet and Kinetics. For the loss, we set $\lambda_{1} = 1 $, $\lambda_{2} = 1 $, and $\lambda_{3} = 2 $. For 10-tasks setup, we train iCaRL for [25, 25, 15] epochs and BiC for [50, 25, 50 ] epochs on [UCF101, Kinetics, ActivityNet], respectively. For 20-tasks setup, we train iCaRL for [25,15,15] epochs and BiC for [60, 20, 55] epochs on [UCF101, Kinetics, ActivityNet], respectively.

\vspace{3pt}\noindent\textbf{Efficient Calculation of the Secondary Loss.}
Since we work with TSN, the calculation of the secondary backward pass does not incur additional computational costs. The TSN framework already makes independent forward passes for a set of frames sampled from the original video, then averages their logits to get the video's prediction. As a result, we can simply obtain the logits of any boring version of the video as a copy of the logits found in the original TSN forward pass. We simply perform an extra loss calculation on the copy of the logits and accumulate the losses in the backward pass.

\subsection{Comparison Against the State-of-the-art}
\textbf{Video CIL on 10-Tasks.} We first evaluate SMILE on the 10-task setup proposed in \cite{villa2022vclimb} using BiC and iCaRL. In Table \ref{table:main results 10}, we report the results for SMILE along with the current state-of-the-art on UCF101, ActivityNet, and Kinetics. We also include the performance of the memory-efficient regime proposed in \cite{villa2022vclimb}. 

On every dataset, SMILE outperforms the current state-of-the-art. For UCF101, SMILE shows a significant improvement in acc over both baselines by 14.73\% (iCaRL) and 14.33\% (BiC). We highlight that SMILE is only marginally below the standard fully supervised training setup~\cite{gowda2021smart} (training on the whole dataset in one stage) by  2.94\%. Our experiments on ActivityNet show that SMILE outperforms baselines with an Acc of 54.83\% for BiC and 50.26\% for iCaRL, improving over the previous state-of-the-art vCLimb (BiC) by 2.87\%. Finally, SMILE shows a significant improvement on the challenging Kinetics dataset, improving by 20.2\% (BiC) over the previous state-of-the-art vCLimb (iCaRL). 

We also note that backward forgetting is significantly reduced. In the case of the UCF101 dataset, SMILE reports BWF of 2.59\% for iCaRL and 2.15\% for BIC. A similar situation is observed in ActivityNet, where the BWF is reduced by 3.85\% for iCaRL and 16.58\% for BiC. Finally, the backward forgetting for Kinetics drops from 38.74\% to 7.34\% for iCaRL and from 51.96\% to 6.25\% for BiC.

Furthermore, \cite{villa2022vclimb} also proposed a memory efficient regime named 'Temporal Consistency' (TC). This regime explored pushing only 4, 8 and 16 frames and applying a feature regularization for sub-sampled video data. Although our proposal relies on single frame memory, SMILE outperforms any memory regime explored in \cite{villa2022vclimb} by at least 19.86\% in UCF101, 9.10\% in ActivityNet and 15.7\% in Kinetics.

In Figure~\ref{figure:fig1}, we provide further analysis of the forgetting of SMILE using iCaRL on UCF101 dataset for the 10 task setup. We show that SMILE suffers far less from forgetting when compared to the baseline approaches that push 4, 8, 16 and 32 frames into the working memory. We also note that there is a significant difference in forgetting as the task set progresses. While the 4 and 8 frame baselines are close to SMILE during the first 4 tasks, they exhibit significant forgetting in the final 4 tasks (tasks 7 to 10). Meanwhile, SMILE suffers only marginally from forgetting in the same task-subset.

\input{Tables/icarl_20}

\textbf{Video CIL on 20-Tasks.} We conclude the state-of-the-art comparison by addressing the more challenging 20-tasks setup proposed in \cite{villa2022vclimb}. We report the performance of SMILE on UCF101, ActivityNet, and Kinetics along with the current state-of-the-art in Table \ref{table:main results 20}. Despite the increased difficulty of the 20-tasks scenario, we observe that SMILE still outperforms the state-of-the-art on all datasets by up to 21.49\% (Kinetics). Our approach outperforms the baselines in every scenario, except when we use iCaRL with ActivityNet where we get nearly on-par results, an Acc of 43.45\% compared to the baseline's Acc of 43.33\%.

\input{figures/fig1}

\subsection{Effect of Replay Memory Size}

After performing the state-of-the-art comparison, we ablate our main design choices and explore their individual contributions. In Table \ref{table:main results 10}, the memory budget for SMILE is defined by the scale of the training dataset (1 frame per video). These  memory sizes are 9280, 15410, and 246282 frames for UCF101, ActivityNet, and Kinetics, respectively. This already represents a reduction of 97.5\%, 99.9\%, and 87.7\% of the memory budget proposed in vCLIMB. In this section, we investigate the performance of SMILE as we reduce the memory budget further.

\input{Tables/Memory_Budget_UCF}
\input{Tables/Memory_Budget_AN}

Tables \ref{table:memeory ucf} and \ref{table:memory budget AN}, summarize the results of running iCaRL on UCF101 and ActivityNet using the 10-tasks setup for smaller memory budgets. For UCF101, we can see that by reducing the memory size from 9280 to 4040 frames, which represents a relative reduction of 56.47\% in memory, we get an Acc of $86.07\%$ which is still higher than the results reported by \cite{villa2022vclimb} ($80.97\%$). Moreover, by setting the memory budget to 2020 frames, which is about 21.77\% of the original memory budget, we are on par with the results of \cite{villa2022vclimb}.
\newline
For ActivityNet, Table \ref{table:memory budget AN} shows that by reducing the memory size from 15410 to 12000 frames which represents a relative reduction of 22.13\% in memory, we get an Acc of 49.10\% compared to \cite{villa2022vclimb} (48.53\%). A further reduction of the memory to 8000 frames will report an Acc of 49.39\%, which is slightly higher than the results of \cite{villa2022vclimb}.

\subsection{Effect of Frame Selection}

It would be expected that, under such extreme sampling circumstances, the selection strategy for memory frames would play a key role in boosting the effectiveness of our approach. We empirically assess that a structured frame selection does not bring noticeable improvements to our approach. Using TSN as our backbone, we simultaneously obtain global video features (for classification) and frame-level features (for the estimation of the global feature) of any frame included in the forward pass. On this basis, we can select the frame whose feature embedding most closely resembles the original video embedding. Intuitively, this will simplify the task of the secondary loss, favoring faster convergence, and will push into the working memory only those frames that contain relevant information for video classification. 

\input{Tables/Frame_Selection}

We explore such an approach and use two metrics to test how similar a frame is to a video in the feature space. Our first metric is the Euclidean distance, where a smaller value indicates a higher similarity. The second metric is cosine similarity, where the larger the value, the higher the similarity. We perform the experiments using such a frame selection strategy over iCaRL on both UCF101 and ActivityNet datasets. Table \ref{table:frame selection} shows that there is no clear benefit in doing this frame selection. Regardless of the dataset or the selected metric, we could only get slight improvement for UCF101 using the cosine similarity. In fact, using this selection criterion for ActivityNet resulted in no improvement over random selection. Based on these observations, we adopted random sampling for frame selection in SMILE.

\input{figures/PCA}
\subsection{Effect of The Secondary Loss}

As outlined in Section \ref{sec:methodology}, we perform a secondary backward pass over the boring video data and supervise it with the original ground truth of the full-length video clip. To conclude the result section, we investigate the effect of this additional cross-entropy loss on the performance of our approach. We analyze the effect of this loss using iCaRL on both UCF101 and ActivityNet, and we assess the effectiveness of SMILE while enabling and disabling this additional loss. The results are summarized in Table \ref{table:CE}.  
\input{Tables/additional_CE}
Overall, the use of this additional loss leads to an improvement of 11.3\% and 3.86\% on UCF101 and ActivityNet, respectively.  

We further explore the effect of including the secondary loss in SMILE by analyzing the PCA projected features of the original videos and the boring videos associated with them. In Figure \ref{figure:fig5}, we show the arrangement of these PCA projections for the original video (shown as an $\star$ in the plot) and the multiple boring videos (shown as circles) that can be built from it. Overall, we see that the plot at the bottom (with secondary cross-entropy) shows a spatial arrangement where the distance between the original video features and the multiple boring videos associated with the same video is smaller than the distance to boring videos of other original video clips. Without the secondary loss, the arrangement of such videos does not seem to follow any pattern.

%% file: Tables/Statistics.tex
\begin{table}[h]
\centering
\renewcommand{\arraystretch}{1.6}

\scalebox{0.8}{
\begin{tabular}{@{}ccccccc@{}}
\toprule
\multirow{2}{*}{\textbf{Dataset}} & \multirow{2}{*}{\textbf{Tasks}} & \multicolumn{3}{c}{\textbf{Videos Per Task}} & \multirow{1}{*}{\textbf{Classes} } & \multirow{1}{*}{\textbf{Avg. Frames}} \\
\cline{3-5}                            &    & \textbf{Train} & \textbf{Val}  & \textbf{Test} &  \textbf{Per Task}  &  \textbf{Per Video}    \\ \midrule
\multirow{2}{*}{\textbf{UCF101}}      & 10 & 928   & 131  & 272  & 10 & 183  \\
                             & 20 & 464   & 65   & 136  & 5  & 183  \\ \midrule
\multirow{2}{*}{\textbf{ActivityNet}} & 10 & 1541  & 765  & –    & 20 & 3879 \\
                             & 20 & 770   & 383  & –    & 10 & 3879 \\ \midrule
\multirow{2}{*}{\textbf{Kinetics}}    & 10 & 24628 & 1988 & 3977 & 40 & 250  \\
                             & 20 & 12314 & 994  & 1988 & 20 & 250  \\ \bottomrule
\end{tabular}}
\caption{\textbf{Tasks Statistics.} We outline the key statistics of the task sets proposed in vCLIMB \cite{villa2022vclimb}. The benchmark proposes two task schedules containing a total of 10-tasks and 20-tasks in every dataset. }

\label{table:4}
\end{table}

%% file: Tables/main_results_10.tex
\begin{table*}
\centering
\renewcommand{\arraystretch}{1.3}
\resizebox{0.75 \textwidth}{!}{
\begin{tabular}{cccccccc}
\specialrule{0.8pt}{3pt}{3pt}
\multirow{2}{*}{Model} &
  Frames &
  \multicolumn{2}{c}{UCF101} &
  \multicolumn{2}{c}{ActivityNet} &
  \multicolumn{2}{c}{Kinetics} \\
  \cline{3-8}
 &
  Per Video &
  Acc $\uparrow$ &
  \multicolumn{1}{c}{BWF $\downarrow$} &
  Acc $\uparrow$ &
  \multicolumn{1}{c}{BWF $\downarrow$} &
  Acc $\uparrow$ &
  \multicolumn{1}{c}{BWF $\downarrow$} \\
   
\specialrule{0.8pt}{3pt}{3pt}

iCaRL+SMILE (Ours)  & 1  & \textbf{95.70} & \textbf{2.59}  & 50.26 & 15.87 &  46.58  & 7.34 \\ \hline
BiC+SMILE (Ours)   & 1   & 92.49 & 2.15 & \textbf{54.83}   &    \textbf{7.69}   &   \textbf{52.24}    &   \textbf{6.25}    \\ \hline
vCLimb (iCaRL)  \cite{villa2022vclimb} & N.A.* & 80.97 & 18.11 & 48.53 & 19.72 & 32.04 & 38.74 \\ \hline
vCLimb (BiC) \cite{villa2022vclimb}  & N.A.* & 78.16 & 18.49 & 51.96 & 24.27 & 27.90 & 51.96 \\ \hline
vCLimb (ICaRL+TC) \cite{villa2022vclimb} & 16  & 75.84 & 23.23 &44.04 & 22.82 & 36.54 & 33.53 \\ \hline
vCLimb (ICaRL+TC) \cite{villa2022vclimb} & 8  & 74.25 & 25.27 & 45.73 & 18.90 & 36.24 & 33.83 \\ \hline
vCLimb (ICaRL+TC) \cite{villa2022vclimb} & 4  & 73.85 & 26.35 & 42.99 & 23.82 & 35.32 & 34.07 \\ \hline

\end{tabular}
}
\caption{\textbf{State-of-the-art comparison vCLIMB (10-tasks).} We report our results on the vCLIMB benchmark, along with the state-of-the-art outlined in \cite{villa2022vclimb}. Despite storing at least 87.7\% less frames in memory, SMILE reports state-of-the-art performance for every dataset included in the vCLIMB benchmark. SMILE also outperforms the low-memory setting of \cite{villa2022vclimb} in every dataset and memory budget.*: N.A. indicates that the number of saved frames per video is not fixed.}
\label{table:main results 10}
\end{table*}

%% file: Tables/icarl_20.tex
\begin{table*}
\centering
\renewcommand{\arraystretch}{1.3}
\resizebox{0.65 \textwidth}{!}{
\begin{tabular}{ccccccc}
\specialrule{0.8pt}{3pt}{3pt}
\multirow{2}{*}{Model} &
  
  \multicolumn{2}{c}{UCF101} &
  \multicolumn{2}{c}{ActivityNet} &
  \multicolumn{2}{c}{Kinetics} \\
  \cline{2-7}
 &
  
  Acc $\uparrow$ &
  \multicolumn{1}{c}{BWF $\downarrow$} &
  Acc $\uparrow$ &
  \multicolumn{1}{c}{BWF $\downarrow$} &
  Acc $\uparrow$ &
  \multicolumn{1}{c}{BWF $\downarrow$} \\

\specialrule{0.8pt}{3pt}{3pt}

iCaRL+SMILE (Ours)    & \textbf{95.09} & \textbf{1.72}  & 43.45 & 23.02 &  45.77  &  4.57\\ \hline

BiC+SMILE (Ours)    & 90.90 & 1.62& \textbf{51.14}& \textbf{1.33}& \textbf{48.22} & \textbf{0.31}\\ \hline

vCLimb (iCaRL)  \cite{villa2022vclimb}  & 76.59           & 21.83             & 43.33             & 21.57 & 26.73 & 42.25 \\ \hline
vCLimb (BiC) \cite{villa2022vclimb} & 70.69           & 24.90             & 46.53             & 15.95& 23.06 & 58.97\\
\hline
\end{tabular}
}
\caption{\textbf{SMILE \vs the state-of-the-art performance (20-tasks).} We report our results along with vCLIMB’s using both average accuracy (Acc)
and backward forgetting (BWF) on the challenging 20-tasks setup. We outperform the state-of-the-art for all datasets.}
\label{table:main results 20}
\end{table*}

%% file: figures/fig1.tex
\begin{figure}[t]
\begin{center}
\includegraphics[width=0.46\textwidth]{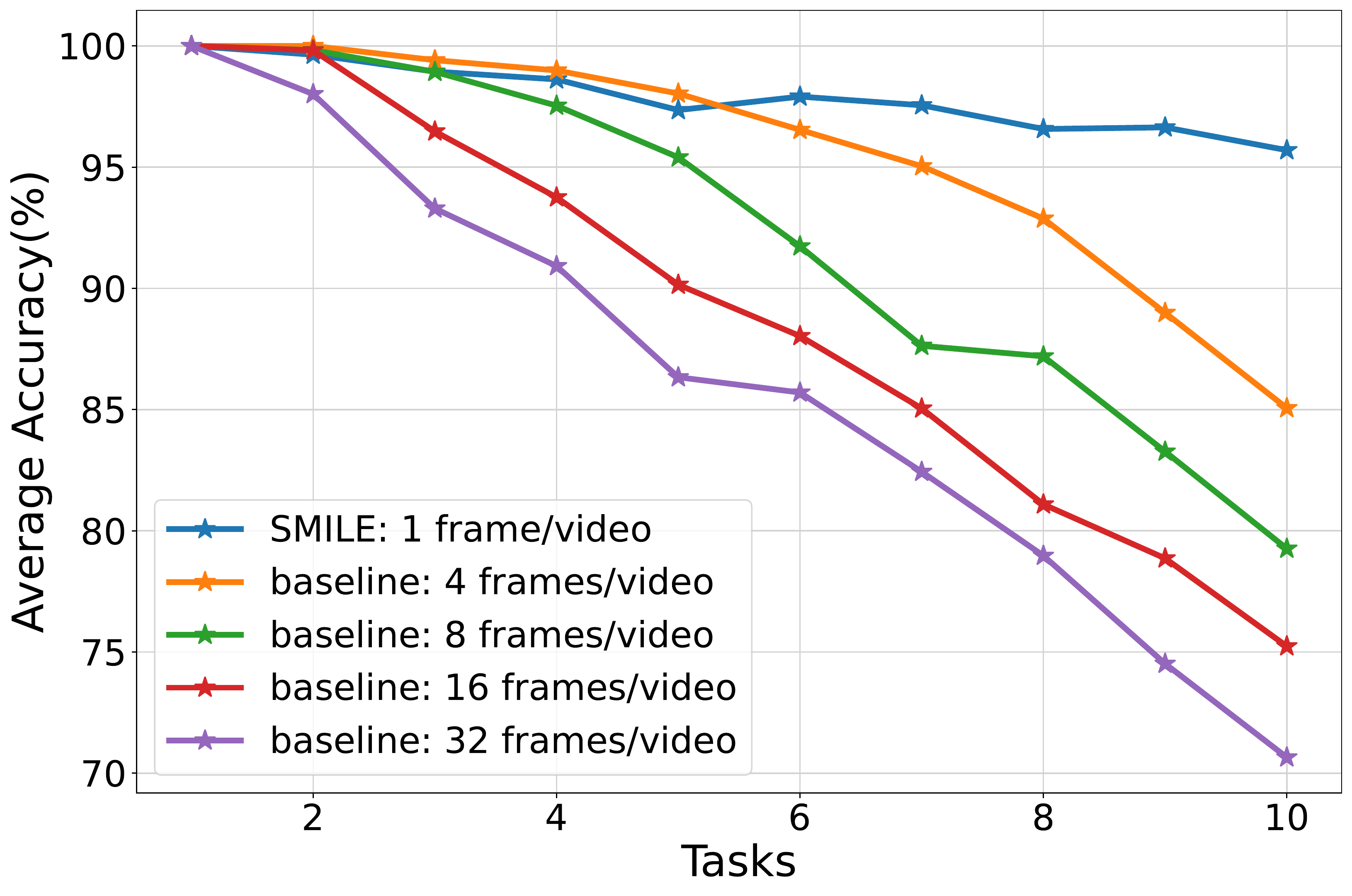}
\small
\caption{\textbf{Average Accuracy for the 10-task setup with different numbers of frames stored in the replay memory compared to iCaRL+SMILE (UCF101).} The trend clearly shows how storing fewer frames in the memory (thus allowing to preserve more videos), leads to performance boosts.}
\label{figure:fig1}
\end{center}
\end{figure}

%% file: Tables/Memory_Budget_UCF.tex
\begin{table}
\centering

\renewcommand{\arraystretch}{1.3}
\begin{tabular}{@{}c
>{\columncolor[HTML]{EFEFEF}}c c
>{\columncolor[HTML]{EFEFEF}}c c@{}}
\specialrule{0.8pt}{0pt}{0.5pt}
\textbf{Memory Size (Frames)} & \textbf{2020}    & \textbf{4040}    & \textbf{8080}   & \textbf{9280} \\ \hhline{-----}
Acc$\uparrow$         & 78.01 & 86.07 & 94.97 &  95.70 \\
BWF$\downarrow$       & 15.68 & 10.36 & 2.32 &  2.59 \\ \specialrule{0.8pt}{0pt}{0.5pt}
\end{tabular}
\caption{\textbf{Effect of memory budget (UCF101)}. We tested even smaller memory budgets for SMILE on UCF101 using iCaRL as CL method. We discover that even with just 2020 total frames (storing a single frame for less than 1/2 of the dataset) SMILE is just below the current state-of-the-art. SMILE actually outperforms the state-of-the-art in every other scenario.}
\label{table:memeory ucf}
\end{table}

%% file: Tables/Memory_Budget_AN.tex
\begin{table}
\centering
\renewcommand{\arraystretch}{1.3}

\begin{tabular}{@{}c
>{\columncolor[HTML]{EFEFEF}}c c
>{\columncolor[HTML]{EFEFEF}}c c@{}}
\specialrule{0.8pt}{0pt}{0.5pt}
\textbf{Memory Size (Frames)} & \textbf{4000}  & \textbf{8000}  & \textbf{12000} & \textbf{15410} \\ \hhline{-----}
Acc$\uparrow$         & 45.04 & 49.39 & 49.10 & 50.26 \\
BWF$\downarrow$         & 20.82 & 17.56 & 18.39 & 15.87 \\ \specialrule{0.8pt}{0pt}{0.5pt}
\end{tabular}
\caption{\textbf{Effect of memory budget (ActivityNet).} We also test small memory regimes for the ActitvityNet dataset using iCaRL. When compared against the iCaRL methods in \cite{villa2022vclimb}, we still outperform their best performance while storing around 1/2 of the dataset.}
\label{table:memory budget AN}
\end{table}

%% file: Tables/Frame_Selection.tex
\begin{table}
\centering
\renewcommand{\arraystretch}{1.2}

\scalebox{0.95}{
\begin{tabular}{@{}cllll@{}}
\toprule
\multirow{2}{*}{\textbf{Frame Selection}} & \multicolumn{2}{c}{\textbf{UCF101}}                        & \multicolumn{2}{c}{\textbf{ActivityNet}}                   \\ \cmidrule(l){2-5} 
                                 & \multicolumn{1}{c}{Acc $\uparrow$} & \multicolumn{1}{c}{BWF$\downarrow$} & \multicolumn{1}{c}{Acc$\uparrow$} & \multicolumn{1}{c}{BWF $\downarrow$} \\ \midrule
Random            & 95.70 & 2.59 & \textbf{50.26} & \textbf{15.87} \\
Euclidean         & 95.34 & 2.37 & 50.25 & 16.37 \\
Cosine Similarity & \textbf{96.25} & \textbf{2.00} & 49.98 & 16.02 \\ \bottomrule
\end{tabular}}

\caption{\textbf{Analyzing Frame Selection Strategies}. We test different frame selection strategies to select a single frame that most closely resembles the full video features and then push it into the memory.}
\label{table:frame selection}
\end{table}

%% file: figures/PCA.tex
\begin{figure}[t]
\begin{center}
\includegraphics[width=0.35\textwidth]{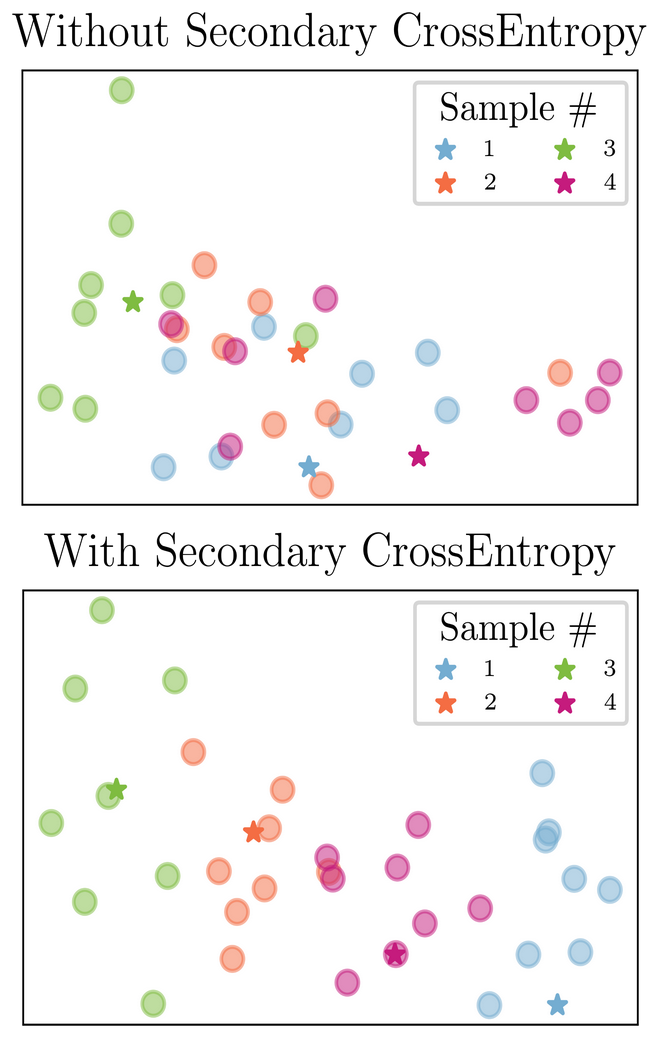}
\small
\caption{\textbf{Projection of Video and Boring Videos Features using PCA.} We show that training using iCaRL on UCF101 with an additional cross-entropy helps us learn a representation where the PCA projected features of a video and its corresponding boring videos are more similar. The projected features of the original video samples are marked with $\star$, whereas those of the boring videos are marked by a circle.}
\label{figure:fig5}
\end{center}
\vspace{-8mm}
\end{figure}

%% file: Tables/additional_CE.tex
\begin{table}[h]
\centering
\scalebox{0.95}{
\begin{tabular}{@{}clrlr@{}}
\toprule
\multirow{2}{*}{\textbf{Secondary Loss}} & \multicolumn{2}{c}{\textbf{UCF101}} & \multicolumn{2}{c}{\textbf{ActivityNet}} \\ \cmidrule(l){2-5} 
          & Acc$\uparrow$                        & \multicolumn{1}{l}{BWF$\downarrow$} & Acc$\uparrow$                        & \multicolumn{1}{l}{BWF$\downarrow$} \\ \midrule
\xmark     & \multicolumn{1}{r}{84.40} & 0.78                  & \multicolumn{1}{r}{46.40} & 9.36                   \\
\cmark & \multicolumn{1}{r}{\textbf{95.70}}  & \textbf{2.59}                    & \multicolumn{1}{r}{\textbf{50.26}} & \textbf{15.87}                   \\ \bottomrule
\end{tabular}}
\caption{\textbf{The Effect of Secondary Loss.} We evaluate the effect of adding the secondary loss using iCaRL on both UCF101 and ActivityNet.}
\label{table:CE}
\end{table}

%% file: sections/conclusion.tex
\section{Conclusion}
\label{sec:conclusion}

In this work, we introduced SMILE, a simple yet effective replay mechanism for video continual learning. We showed that, by keeping a single video frame in memory, state-of-the-art results can be achieved in every dataset of the challenging vCLIMB benchmark. Despite reducing the working memory size by at least 87.7\%, SMILE outperforms every other video CIL in the 10 and 20 task schedules proposed in vCLIMB. Remarkably, SMILE achieves this performance by using standard image CIL methods in the video domain. This is possible as SMILE's single frame memory enables the direct use of image CIL methods in the video domain. 
\newline
\newline
\noindent\textbf{Acknowledgments.} This work is supported by the King Abdullah University of Science and Technology (KAUST) Office of Sponsored Research (OSR) under Award No. OSR-CRG2021-4648, as well as, the SDAIA-KAUST Center of Excellence in Data Science and Artificial Intelligence (SDAIA-KAUST AI).